\documentclass[sn-mathphys,Numbered,iicol]{sn-jnl}%

\usepackage{graphicx}%
\usepackage{multirow}%
\usepackage{mathbbol}
\usepackage{amsmath,amssymb,amsfonts,bm}%
\usepackage{amsthm}%
\usepackage{mathrsfs}%
\usepackage[title]{appendix}%
\usepackage{xcolor}%
\usepackage{textcomp}%
\usepackage{manyfoot}%
\usepackage{booktabs}%
\usepackage{algpseudocode}
\usepackage{algpseudocode}%
\usepackage{listings}%
\usepackage{subcaption} 
\usepackage{diagbox}
\usepackage[ruled,vlined]{algorithm2e}

\theoremstyle{thmstyleone}%

\theoremstyle{thmstyletwo}%

\theoremstyle{thmstylethree}%

\begin{document}

\newcommand{\fracpartial}[2]{\frac{\partial #1}{\partial  #2}}
\newcommand{\norm}[1]{\left\lVert#1\right\rVert}
\newcommand{\innerproduct}[2]{\left\langle#1, #2\right\rangle}
\newcommand{\fan}[1]{\Vert #1 \Vert}
\newcommand{\qileft}{[\kern-0.15em[}
\newcommand{\qiLeft}{\left[\kern-0.4em\left[}
\newcommand{\qiright}{]\kern-0.15em]}
\newcommand{\qiRight}{\right]\kern-0.4em\right]}
\newcommand{\sign}{{\mbox{sign}}}
\newcommand{\diag}{{\mbox{diag}}}
\newcommand{\armin}{{\mbox{argmin}}}
\newcommand{\rank}{{\mbox{rank}}}
\renewcommand{\vec}{{\mbox{vec}}}
\newcommand{\st}{{\mbox{s.t.}}}
\newcommand{\<}{\left\langle}
\renewcommand{\>}{\right\rangle}
\newcommand{\lbar}{\left\|}
\newcommand{\rbar}{\right\|}
\newcommand{\eg}{{\emph{e.g.}}}
\newcommand{\ie}{{\emph{i.e.}}}
\newcommand{\wrt}{{{w.r.t.}}}
\renewcommand{\Roman}[1]{\uppercase\expandafter{\romannumeral#1}}
\newcommand{\red}[1]{{\color{red}{#1}}}
\newcommand{\blue}[1]{{\color{blue}{#1}}}

\renewcommand{\a}{{\bm{a}}}
\renewcommand{\b}{{\bm{b}}}
\renewcommand{\d}{{\bm{d}}}
\newcommand{\e}{{\bm{e}}}
\newcommand{\f}{{\bm{f}}}
\newcommand{\g}{{\bm{g}}}
\renewcommand{\o}{{\bm{o}}}
\newcommand{\p}{{\bm{p}}}
\newcommand{\q}{{\bm{q}}}
\renewcommand{\r}{{\bm{r}}}
\newcommand{\s}{{\bm{s}}}
\renewcommand{\t}{{\bm{t}}}
\renewcommand{\u}{{\bm{u}}}
\renewcommand{\v}{{\bm{v}}}
\newcommand{\w}{{\bm{w}}}
\newcommand{\x}{{\bm{x}}}
\newcommand{\y}{{\bm{y}}}
\newcommand{\z}{{\bm{z}}}
\newcommand{\balpha}{{\bm{\alpha}}}
\newcommand{\bbeta}{{\bm{\beta}}}
\newcommand{\bmu}{{\bm{\mu}}}
\newcommand{\bsigma}{{\bm{\sigma}}}
\newcommand{\blambda}{{\bm{\lambda}}}
\newcommand{\btheta}{{\bm{\theta}}}
\newcommand{\bgamma}{{\bm{\gamma}}}
\newcommand{\bxi}{{\bm{\xi}}}
\newcommand{\bphi}{{\bm{\phi}}}

\newcommand{\ba}{{\bm{A}}}
\newcommand{\bb}{{\bm{B}}}
\newcommand{\bc}{{\bm{C}}}
\newcommand{\bd}{{\bm{D}}}
\newcommand{\be}{{\bm{E}}}
\newcommand{\bg}{{\bm{G}}}
\newcommand{\bi}{{\bm{I}}}
\newcommand{\bj}{{\bm{J}}}
\newcommand{\bl}{{\bm{L}}}
\newcommand{\bo}{{\bm{O}}}
\newcommand{\bp}{{\bm{P}}}
\newcommand{\bq}{{\bm{Q}}}
\newcommand{\bs}{{\bm{S}}}
\newcommand{\bu}{{\bm{U}}}
\newcommand{\bv}{{\bm{V}}}
\newcommand{\bw}{{\bm{W}}}
\newcommand{\bx}{{\bm{X}}}
\newcommand{\by}{{\bm{Y}}}
\newcommand{\bz}{{\bm{Z}}}
\newcommand{\bTheta}{{\bm{\Theta}}}
\newcommand{\bSigma}{{\bm{\Sigma}}}

\newcommand{\A}{{\mathcal{A}}}
\newcommand{\B}{\mathcal{B}}
\newcommand{\C}{\mathcal{C}}
\newcommand{\D}{\mathcal{D}}
\newcommand{\F}{\mathcal{F}}
\renewcommand{\H}{\mathcal{H}}
\newcommand{\I}{\mathcal{I}}
\renewcommand{\L}{\mathcal{L}}
\newcommand{\N}{\mathcal{N}}
\renewcommand{\P}{\mathcal{P}}
\newcommand{\X}{\mathcal{X}}
\newcommand{\Y}{\mathcal{Y}}
\newcommand{\W}{\mathcal{W}}

\title[CoNe: Contrast Your Neighbours for Supervised Image Classification]{CoNe: Contrast Your Neighbours for Supervised Image Classification}

\author[1]{\fnm{Mingkai} \sur{Zheng}}\email{mingkaizheng@outlook.com}

\author*[2]{\fnm{Shan} \sur{You}}\email{youshan@sensetime.com}

\author[3]{\fnm{Lang} \sur{Huang}}\email{langhuang@cvm.t.u-tokyo.ac.jp}

\author[1]{\fnm{Xiu} \sur{Su}}\email{xisu5992@uni.sydney.edu.au}

\author[4]{\fnm{Fei} \sur{Wang}}\email{wangfei91@mail.ustc.edu.cn}

\author[2]{\fnm{Chen} \sur{Qian}}\email{qianchen@sensetime.com}

\author[5]{\fnm{Xiaogang} \sur{Wang}}\email{xgwang@ee.cuhk.com.hk}

\author[1]{\fnm{Chang} \sur{Xu}}\email{c.xu@sydney.edu.au}

\affil[1]{\orgname{The University of Sydney}}
\affil[2]{\orgname{SenseTime Research}}

\affil[3]{\orgname{The University of Tokyo}}
\affil[4]{\orgname{University of Science and Technology of China}}
\affil[5]{\orgname{The Chinese University of Hong Kong}}

\abstract{Image classification is a longstanding problem in computer vision and machine learning research. Most recent works (\eg~SupCon \cite{SupervisedCL}, Triplet \cite{tripletloss}, and max-margin \cite{max-margin}) mainly focus on grouping the intra-class samples aggressively and compactly, with the assumption that all intra-class samples should be pulled tightly towards their class centers. However, such an objective will be very hard to achieve since it ignores the intra-class variance in the dataset. (\ie~different instances from the same class can have significant differences). Thus, such a monotonous objective is not sufficient. To provide a more informative objective, we introduce Contrast Your Neighbours (CoNe) - a simple yet practical learning framework for supervised image classification. Specifically, in CoNe, each sample is not only supervised by its class center but also directly employs the features of its similar neighbors as anchors to generate more adaptive and refined targets. Moreover, to further boost the performance, we propose ``distributional consistency" as a more informative regularization to enable similar instances to have a similar probability distribution. Extensive experimental results demonstrate that CoNe achieves state-of-the-art performance across different benchmark datasets, network architectures, and settings. Notably, even without a complicated training recipe, our CoNe achieves 80.8\% Top-1 accuracy on ImageNet with ResNet-50, which surpasses the recent Timm training recipe \cite{timmresnet} (80.4\%). Code and pre-trained models are available at \href{https://github.com/mingkai-zheng/CoNe}{https://github.com/mingkai-zheng/CoNe}.}


\maketitle

\begin{figure*}[!t]
    \centering
    \includegraphics[width=0.7\linewidth]{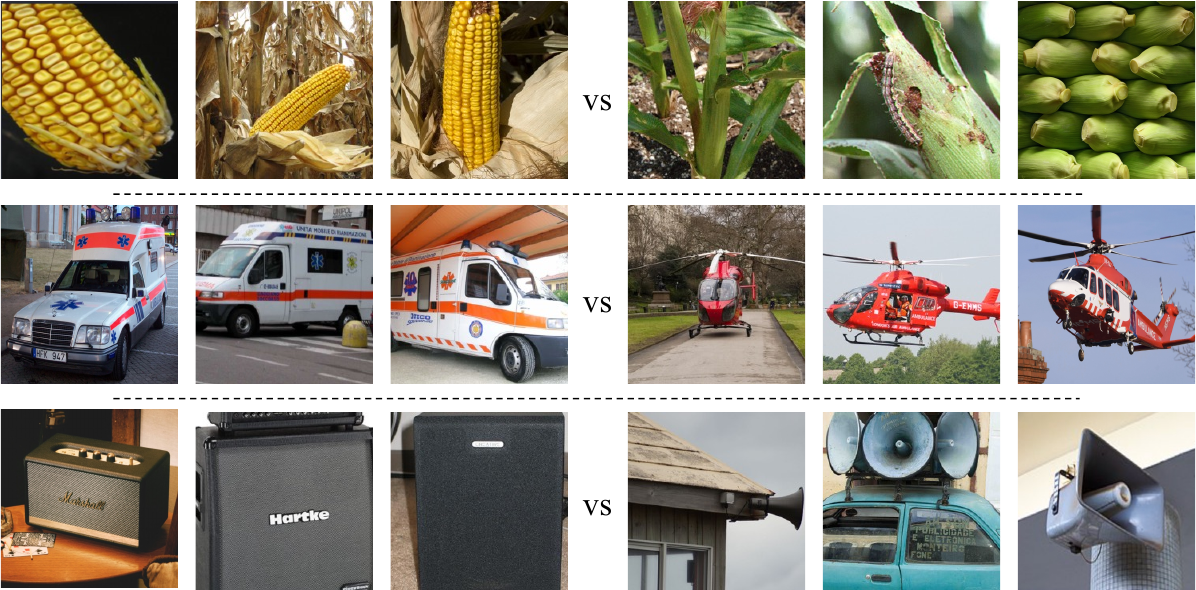}
    \caption{Examples to show the high variance among intra-class samples. These images are all from ImageNet class. Images in the same row belong to the same class. $1^{st}$ row: peeled corn \emph{vs.} unpeeled corn. $2^{ed}$ row: ambulance car \emph{vs.} ambulance helicopter. $3^{rd}$ row: different types of speakers.  }
    \label{fig:vs}
\end{figure*}

\section{Introduction}
Deep neural networks have demonstrated superior performance on various visual tasks \cite{imagenet_cvpr09, pascal-voc-2007, coco, cifar}. In particular, image classification is considered the most fundamental task because of its simplicity and various real-world applications. It also serves as a kind of pre-training task since the learned representations can be easily transferred into various downstream tasks (\eg~object detection, video analysis, and semantic segmentation). Thus, improving the image classification performance has become a longstanding problem; an enormous number of regularization methods \cite{mixup, cutmix, dropout, dropblock, inception, autoaugment, randaugment, stochasticdepth} and training strategies \cite{accurate, SupervisedCL, max-margin} have been proposed to address this issue.

A typical image classification algorithm maintains a set of class centers. The training objective aims to maximize the inner product of the feature vectors with their corresponding class center while minimizing the inner product with respect to other class centers. Previous feature learning based methods \eg max-margin \cite{max-margin} directly modified the logits calculation, which aims to decouple the magnitude and direction of the feature vectors, and shows that the learned feature could be more discriminative as we maximize and minimize the cosine similarity for the inter-class and intra-class samples. Many works \cite{CosFace, ArcFace, NormFace, SphereFace, CircleLoss} extend this idea and show their promising performance in various domains (\eg, Facial Recognition and Person Re-identification). More recently, thanks to the success of contrastive learning \cite{simclr, moco, byol, barlowtwins, swav, ressl, WCL},  SupCon has been proposed to bring the InfoNCE loss into a fully supervised setting by allowing an arbitrary number of positives samples in the loss functions. Theoretical analysis shows that the $\mathcal{L}^{sup}_{out}$ in SupCon behaves similarly to the triplet loss \cite{triplet} and hard positive mining.

The objective of these methods are the same during training, which is to group the intra-class samples aggressively and compactly, with the assumption that all intra-class samples should be pulled tightly toward their class centers. However, such an objective is tough to achieve and against the intrinsic characteristics, especially for those datasets (\eg~ImageNet \cite{imagenet_cvpr09}) with high intra-class variance. As we have shown some examples in Figure \ref{fig:vs}, we can observe that different instances from the same class can have significant differences. In this case, a monotonous class center might lose its adaptivity to cover this intra-class variance for more intrinsic training. 

Intuitively, instead of using a single class center as the target for the intra-class samples, those similar positives should also be good targets since such targets are semantically and visually similar. Thus, to provide more informative supervision, we would like to explicitly apply an additional constraint to let the similar positives close in the embedding space. Fortunately, we found that the $\mathcal{L}^{sup}_{in}$ has exactly the properties we need, although the original SupCon paper claims that it fails to learn a good representation. In this paper, we show that the underlying property of $\mathcal{L}^{sup}_{in}$ is to encourage the features of training samples to be pulled towards their similar positives adaptively. The reason for the failure of $\mathcal{L}^{sup}_{in}$ results from the ignorance of the hard positive samples. Based on this observation, we thus propose a new supervised learning framework - Contrast Your Neighbors (CoNe), which utilizes the $\mathcal{L}^{sup}_{in}$ more appropriately. Concretely, we simply take the classical cross-entropy loss to ensure the intra-class samples are constrained to have the same targets, and then we can arm the $\mathcal{L}^{sup}_{in}$ with the nearest neighbors of the training features to construct a more semantic aware and refined targets to help the compactness of the intra-class samples. Furthermore, we also propose ``distributional consistency" regularization to enhance the compactness between similar samples by encouraging features with similar probability distribution as their neighbors. Extensive experimental results on multiple settings show the superiority of CoNe. Our contributions can be summarized as follows.

\begin{itemize}
    \item We propose a novel supervised learning framework (CoNe) and utilize the $\mathcal{L}^{sup}_{in}$ more properly for the image classification task.
    
    \item  We theoretically analyze the $\mathcal{L}^{sup}_{in}$ loss and show that the training sample features will be pulled towards its similar positive features, and more similar features will have greater contribution to the gradients.
    
    \item  We propose a novel distributional consistency regularization that encourages similar samples to predict similar probability distributions to improve the performance further.
    
    \item Our proposed method achieves state-of-the-arts performance for image classification tasks. Experiments results shows that our method has better performance than Timm training recipe \cite{timmresnet}. For example, with ResNet-50 as the backbone network, CoNe achieves 80.8\% Top-1 accuracy on ImageNet.
\end{itemize}

\section{Related work}
\subsection{Training Strategies}
Many methods have been proposed to improve image classification accuracy. Most of these methods aim to make the training process harder to prevent the network from overfitting. For example, \cite{autoaugment, randaugment, fastaugment} propose a set of rich data augmentation policies to expand the variety of the datasets. Some regional dropout based methods \cite{cutout, erase} show that removing random regions in images is a simple but effective regularization that improves the classification performance. DropBlock \cite{dropblock} further generalizes this idea by removing a random block from the feature map. Stochastic Depth \cite{stochasticdepth} is a similar idea that randomly drops the entire layer during training. Mixup \cite{mixup} adopts a convex combination of pairs of images and uses the corresponding mixed labels as the targets, improving the classification accuracy and robustness of the adversarial samples. CutMix \cite{cutmix} combines the idea of Mixup and Cutout, which randomly mix two patches from a pair of images and generate the targets by linear interpolating the one-hot labels based on the area of the two patches. Recent work \cite{timmresnet} has explored the optimal combination of these training strategies and significantly improved the baseline results.

\subsection{Representation Learning.}
Recently, because of the success of the contrastive learning \cite{moco, mocov2, mocov3, simclr, SimSiam, simclrv2, goodview, swav, byol, nnclr, adco, ressl, resslv2, simmatch, simmatchv2, comatch, WCL, prcl, lewel}, SupCon has been proposed to extend the idea of contrastive learning to the fully-supervised setting, which allows the InfoNCE objective to be applied with an arbitrary number of positives. Experimental results in SupCon show that it improves the performance and robustness of the image classification problem. More recently, some works \cite{sl-mlp} have started rethinking the performance gaps between the supervised and unsupervised pre-training and improving the transferability of the supervised pre-trained model by adding a simple MLP projector. In this paper, we show that CoNe can enhance the performance of both upstream (classification) and downstream tasks  (detection, segmentation, and downstream classification tasks).

\subsection{Deep Metric Learning}
The contrastive loss is not only widely used in recent self-supervised learning, but it is also closely related to deep metric learning (DML). Most of the work in DML focuses on how to use hard/easy negatives/ positives to form a better triplet. For example, triplet loss \cite{tripletloss, triplet_sampling} suggests that semi-hard negatives are better than hard negatives. \cite{defense_triplet} propose a soft margin for triplet loss and use all possible negatives in a batch to omit the complex sampling tricks. Meanwhile, many DML papers \cite{CosFace, ArcFace, CircleLoss, amsoftmax} also focus on maximizing inter-class margins to get more compact clusters, especially for face verification tasks. In contrast, some works \cite{easy_triplet, Magnet, softtriplet} shows that keeping intra-class variance helps learn better features. Note that keeping intra-class variance features is somewhat similar to this paper. However, different from \cite{easy_triplet, Magnet, softtriplet} proposed for the DML tasks, our CoNe is specially designed for classification problems, which ensures the compactness of the positive samples while keeping the intra-class variance. On the other hand, our CoNe is simpler and more adaptive than \cite{easy_triplet, Magnet} since we do not have the sophisticated design to assign the intra-class samples to different class centers. Most importantly, our CoNe significantly improves the classification performance on the large-scale ImageNet \cite{imagenet_cvpr09} dataset.

\begin{figure*}[t]
    \centering
    \includegraphics[width=\linewidth]{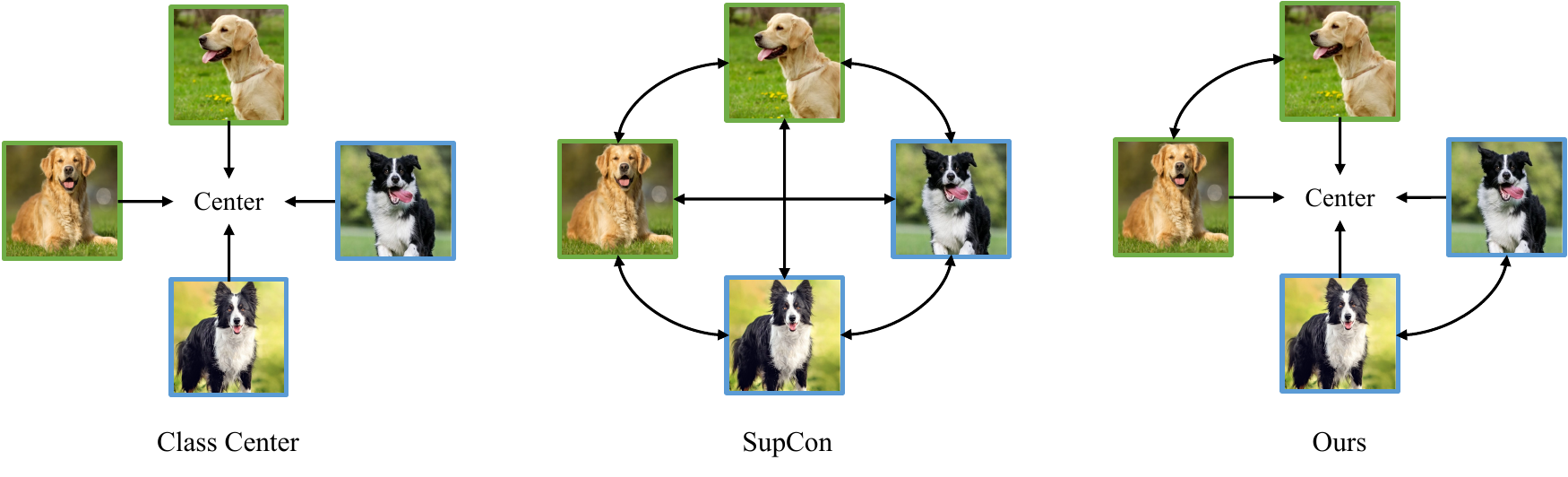}
    \caption{\textbf{Single Arrow}: One sample is pulled towards another. \textbf{Double Arrow}: Two samples are pulled towards each other. \textbf{Left}: Typical class center based methods where all the intra-class samples are pulled towards the same targets. \textbf{Middle}: Supervised Contrastive learning ($\mathcal{L}^{sup}_{out}$) where the features of the samples are directly pulled towards all the positives. Note that the analysis from shows that the harder positive will result in a larger gradient. \textbf{Right}: Our proposed method. We adopt the class center to ensure all intra-class samples can be grouped together, and we also encourage the feature to be pulled toward similar samples.}
    \label{fig:framework}
\end{figure*}

\section{Methodology}
In this section, we will first revisit the problem formulation for image classification; then, theoretically analyze the underlying property of $\mathcal{L}^{sup}_{in}$ and introduce our proposed method CoNe. After that, the algorithm and the implementation details will also be explained.

\subsection{Revisiting Supervised Image Classification}
We define the image classification problem as the following. Given a batch of samples $\bm{x}$, we adopt a convolutional based encoder $\mathcal{F}(\cdot)$ to obtain the corresponding representation vector \ie~$\bm{z} = \mathcal{F}(\bm{x})$. Let us define the $i$-th input sample $\bm{x}_i$ has the label $\bm{y}_i$, then the classical SoftMax with cross-entropy loss could be written as Eq \eqref{equation:ce}, where $W$ is the classifier matrix, $W_c$ is the corresponding class center for class $c$, and $C$ is the total number of classes. 
\begin{equation}
\label{equation:ce}
    \mathcal{L}_{ce} = - \log \frac{\exp({W}_{y_{i}}^{T}  \cdot \bm{z}_i ) }{ \displaystyle\sum_{c=1}^{C} \exp(W_{c}^{T}  \cdot \bm{z}_i )}
\end{equation}
SupCon follows the standard contrastive learning approach where each instance is augmented two times, and both augmented views will be passed into the network to obtain the feature vectors. Specifically, the $\mathcal{L}^{sup}_{in}$ in SupCon can be expressed by Eq. \eqref{equation:supin}, where $\bm{z}$ is the normalized feature. $Pos(i)$ and $Neg(i)$ refer to the anchors that have the same and different ground truth labels. $\tau$ is the temperature parameter that controls the sharpness of the distribution.
\begin{equation}
\label{equation:supin}
\footnotesize{
    \mathcal{L}^{sup}_{in} =  - \log \frac{\displaystyle\sum_{p \in Pos(i)}  \exp(\bm{z}_{i} \cdot \bm{z}_{p} / \tau) }{\displaystyle\sum_{p \in Pos(i)}  \exp(\bm{z}_{i} \cdot \bm{z}_{p} / \tau) + \displaystyle\sum_{n \in Neg(i)}  \exp(\bm{z}_{i} \cdot \bm{z}_{n} / \tau)}
}
\end{equation}
\subsection{Analysis of SupCon Objective}
To better understand the intrinsic property of Eq. \eqref{equation:supin}, we reformulate the equation as the following form Eq. \eqref{equation:objective}, where $\max(\bm{z}_{i} \cdot \bm{z}_{n})$ is the hardest negative, $\max(\bm{z}_{i} \cdot \bm{z}_{p})$ is the most similar positive, and $m$ is the approximation bias for $LogSumExp$ to $\max$ operator. As $\tau \rightarrow 0$, we have $m \rightarrow 0$. However, a commonly used value for $\tau$ is normally from $0.07$ to $0.2$ \cite{simclr, moco, SupervisedCL}, which means Eq. \eqref{equation:objective} is not a strict and rigorous derivation. When $\tau$ is fixed, the value of $m$ is mainly affected by the number of anchors that are close to $\bm{z}_{i}$. In other words, the approximation results of $LogSumExp$ is affected by the most similar (\ie $Top1$ or $max$) anchor or $TopK$ similar anchors of $\bm{z}_{i}$. Thus, the training objective of Eq. \eqref{equation:supin} can be interpreted as maximizing the similarity of the easy positives and minimizing the similarity of the hard negatives. Noted that $max$ operator in Eq. \eqref{equation:objective} is just for simplicity; it does not affect our interpretation even when $m$ is relatively larger. Please see more discussions in our supplementary materials. 
\begin{equation}
\label{equation:objective}
\begin{aligned}
    \mathcal{L}^{sup}_{in} &= \log \bigg[ 1 + \exp(\log \sum_{n \in Neg(i)} \exp(\bm{z}_{i} \cdot \bm{z}_{n} / \tau) \\ & \;\;\;\;\;\;\;\;\;\;\;\;\;\;\;\;\;\;\;\;\;\;\;\;\;\;\;\;\;\;\; - \log \sum_{p \in Pos(i)} \exp(\bm{z}_{i} \cdot \bm{z}_{p} / \tau) ) \bigg] \\
     & \approx \log \bigg[ 1 + \exp( \underbrace{\max(\bm{z}_{i} \cdot \bm{z}_{n} ) - \max(\bm{z}_{i} \cdot \bm{z}_{p} ) + m }_{\text{Objective} } ) \bigg]
\end{aligned}
\end{equation}
\textbf{Gradient Perspective}. Next, we would like to analyze $\mathcal{L}^{sup}_{in}$ from the gradient perspective to show that similar positive anchors will have a more significant contribution to the optimization direction. We first derive the gradient with respect to $\bm{z}_i$ as the following form Eq. \eqref{equation:gradeint} (See more details in supplementary materials). Please note that we ignore the gradient for the negative samples since it does not affect our conclusion (See more details in Appendix B)
\begin{equation}
\label{equation:gradeint}
\footnotesize{
 - \frac{ \partial \mathcal{L}^{sup}_{in} }{ \partial \, \bm{z}_i } \bigg|_{Pos(i)}  = \displaystyle\sum_{p \in Pos(i)} \bm{z}_p \bigg[ \underbrace{ \frac{\exp(\bm{z}_i \cdot \bm{z}_p / \tau) }{S_p} - \frac{\exp(\bm{z}_i \cdot \bm{z}_p / \tau )}{S_p + S_n}}_{\text{Coefficient $\alpha_p$}}  \bigg]
}
\end{equation}
We define $\mathcal{S}_p$ and $\mathcal{S}_n$ as the summation of the exponential cosine similarity for positive sets and negative sets, which can be expressed as
\begin{equation}
\label{equation:gradeint_notes}
\begin{aligned}
    & \mathcal{S}_p  = \sum_{p \in Pos(i)} \exp(\bm{z}_{i} \cdot \bm{z}_{p} / \tau ) \\
    & \mathcal{S}_n  = \sum_{n \in Neg(i)} \exp(\bm{z}_{i} \cdot \bm{z}_{n} / \tau)
\end{aligned}
\end{equation}

Suppose we have two positive samples $\bm{z}_a$, $\bm{z}_b$, and $\bm{z}_i \cdot \bm{z}_a > \bm{z}_i \cdot \bm{z}_b$. To compare the coefficient value, we simply take a subtraction and formulate the equation as Eq. \eqref{equation:compare}
\begin{equation}
\label{equation:compare}
\begin{aligned}
     \alpha_a - \alpha_b  &=  \frac{\exp(\bm{z}_i \cdot \bm{z}_a /  \tau) - \exp(\bm{z}_i \cdot \bm{z}_b / \tau ) }{S_p} \\ & \;\;\;\;\;\;\;\;\;\;\;\;\;\; - \frac{\exp(\bm{z}_i \cdot \bm{z}_a / \tau)   - \exp(\bm{z}_i \cdot \bm{z}_b / \tau)}{S_p + S_n}
\end{aligned}
\end{equation}
Since $\bm{z}_i \cdot \bm{z}_a - \bm{z}_i \cdot \bm{z}_b > 0$, and $S_p > 0, S_n > 0$. Thus, $\alpha_a - \alpha_b$ is always greater than 0, which yields $\alpha_a > \alpha_b$, shows that a more similar positive anchor could have a more significant contribution to the gradient direction, and $\bm{z}_i$ tends to be pulled towards more similar positive anchors. 

The experiments in show that a standalone $\mathcal{L}^{sup}_{in}$ is failed to learn a good representation. The problem can be revealed by Eq. \eqref{equation:objective} and Eq. \eqref{equation:compare}. Basically, Eq. \eqref{equation:compare} shows that the feature $\bm{z}$ will tend to optimize towards more similar positives. Thus, once $\bm{z}$ is close to one or some of the positive features, the objective Eq. \eqref{equation:objective} can be satisfied, and the loss $\mathcal{L}^{sup}_{in}$ will stop optimizing. In other words, a standalone $\mathcal{L}^{sup}_{in}$ ignores the hard positives and does not encourage all intra-class samples will be grouped together since it does not require all $\bm{z}_i \cdot \bm{z}_p$ to be greater than all $\bm{z}_i \cdot \bm{z}_n$.

\subsection{Contrast Your Neighbours}
Although a standalone $\mathcal{L}^{sup}_{in}$ fails to learn decent representations, the underlying property can be incorporated with the class center based methods to improve the compactness of the intra-class features, as
\begin{equation}
\label{equation:cone}
     \mathcal{L}_{CoNe} = \mathcal{L}_{ce} + \lambda_{sup} \mathcal{L}^{sup}_{in}.
\end{equation}
A typical class center based framework aims to pull the intra-class samples towards the same targets. However, such constraint is too restricted since all the intra-class samples are treated equally; the discrepancy among different instances is ignored. $\mathcal{L}^{sup}_{in}$ could adaptively find more similar positives (a.k.a. neighbors) and utilize them as more semantically aware and refined targets to help the optimization. We can thus leverage the nearest neighbors as anchor sets for the loss $\mathcal{L}^{sup}_{in}$. In this way, the Eq. \eqref{equation:ce} ensures all intra-class samples will be pulled together, and $\mathcal{L}^{sup}_{in}$ behaves as an auxiliary term that adaptively pulls the features towards its similar positives as well (as in Figure \ref{fig:framework}). Compared with SupCon, our CoNe has the following two major differences.

1. Our CoNe can be trained end-to-end since the class centers can be directly used for inference, whereas SupCon requires two-stage (pre-training and fine-tuning) to obtain the final model.

2. SupCon follows the standard contrastive learning paradigm where each instance is augmented with two views (\ie~$\bm{x}^{1}$ and $\bm{x}^{2}$) to ensure that each sample will have at least one positive pair in the batch. In this case, $\bm{z}^{2}$ is very likely to become the most similar feature of $\bm{z}^{1}$ since they are just two different views of the same instance. By following Eq. \eqref{equation:objective}, the objective for $\bm{z}^{1}$ is approximately equal to minimize $\max(\bm{z}^{1} \cdot \bm{z}_n) - \bm{z}^{1} \cdot \bm{z}^{2}$. However, since $\bm{z}^{1} \cdot \bm{z}^{2}$ is very likely to be a relatively large number, the objective Eq. \eqref{equation:objective} can be easily achieved and $\mathcal{L}^{sup}_{in}$ will have a low contribution to the optimization. In our implementation, we do not include contrastive views and discard the loss term if the positive anchor does not exist. Using the rich augmentation policies \cite{autoaugment, randaugment, simclr} to generate the contrastive view might help alleviate the problem. However, we show that excluding the contrastive view already works well under the basic augmentation policy (See more details in the ablation study).

\newlength{\textfloatsepsave}\setlength{\textfloatsepsave}{\textfloatsep} \setlength{\textfloatsep}{0pt}
\begin{algorithm*}
\SetAlgoLined
\SetKwInOut{Input}{Input}
\Input{
$\bm{x} $: a batch of samples. y : the ground truth labels $\mathcal{F}_q$ and $\mathcal{F}_k$: The latest and EMA network. $Q_z$ and $Q_p$: the memory bank for storing past features and past class probability distribution. $\bm{W}^q$ and $\bm{W}^k$ : the latest and EMA class centers. }
\While{network not converge} {
    
    $\bm{z}^q = \mathcal{F}_q(x)$

    $\bm{z}^k = \mathcal{F}_k(x)$
    
    $p^{class-q}$ = SoftMax($\bm{W}^{qT} \bm{z}^q)$
    
    $p^{class-k}$ = SoftMax($\bm{W}^{kT} \bm{z}^k$) 
    
    $p^{instance}$ = SoftMax($\bm{z}^k Q_z / \tau$)
    
    $p^{dc}$ = $p^{instance} \cdot p^{class-k}$ 
    
    Calculate loss $\mathcal{L}_{ce}$ by $p^{class-q}$ and y \tcp*{ Eq. \eqref{equation:ce}}
    
    Calculate loss $\mathcal{L}^{sup}_{in}$ by $\bm{z}^q$ and $Q_z$ \tcp*{ Eq. \eqref{equation:supin}}
    
    Calculate loss $\mathcal{L}_{dc}$ by $p^{class-q}$ and $p^{dc}$ \tcp*{ Eq. \eqref{equation:dcloss}}
    
    Update $\mathcal{F}_q$ and $\bm{W}^q$ with loss $\mathcal{L}_{overall}$ \tcp*{ Eq. \eqref{equation:overall}}
    
    Update $\mathcal{F}_k$ and $\bm{W}^k$ by EMA \tcp*{ Eq. \eqref{equation:ema}}
    
    Update the $Q_z$ and $Q_p$ by $\bm{z}^k$ and $p^{class-k}$;
}
\SetKwInOut{Output}{Output}
\Output{The well trained model $\mathcal{F}_q$ and $\bm{W}^q$}
\caption{Contrast with Your Neighbors (CoNe) }
\label{alg:CoNe}
\end{algorithm*}
\setlength{\textfloatsep}{\textfloatsepsave}

\textbf{Momentum Update}. Since $\mathcal{L}^{sup}_{in}$ heavily relies on the anchor points, we need to provide a large number of features to get high-quality anchors. Inspired by \cite{moco, mocov2}, we utilize an exponential moving averaged (EMA) network and maintain a large memory buffer to store $K$ features from the past batch. If we denote $\mathcal{F}_q$ as the latest network and $\mathcal{F}_k$ as the EMA network, the update rule for $\mathcal{F}_k$ can be expressed by Eq. \eqref{equation:ema}, where $m$ is momentum coefficient. All the positive and negative sets are from the memory bank in our implementation.
\begin{equation}
\label{equation:ema}
     \mathcal{F}_k  \leftarrow m \mathcal{F}_k + (1 - m)\mathcal{F}_q
\end{equation}

\textbf{Distributional Consistency.} As we have shown that $\mathcal{L}^{in}_{sup}$ is a feature-level constraint that encourages similar samples to have similar features. To further enhance this property, we would like to generalize this from the perspective of the probability distribution. \ie~ We also encourage similar samples to have similar class predictions. To this end, we introduce the ``distributional consistency" regularization, which encourages the features to have similar predicted probability distribution as their similar anchors. Concretely, the probability of $j^{th}$ anchor in the memory buffer belongs to the class $m$ can be written as Eq. \eqref{equation:semantic_similarity} where $p^{class} \in R^{K\times C}$.
\begin{equation}
\label{equation:semantic_similarity}
    p^{class}_{j, m} = \frac{\exp(W_{m}^{T}  \cdot \bm{z}_j ) }{ \sum_{c=1}^{C} \exp(W_{c}^{T}  \cdot \bm{z}_j ) } 
\end{equation}
The similarity between $\bm{z}_i$ in the current batch with respect to the $j^{th}$ feature in the memory bank can be expressed by Eq. \eqref{equation:instance_similarity} where $p^{instance}_{i} \in R^{1\times K}$. 
\begin{equation}
\label{equation:instance_similarity}
    p^{instance}_{i, j} = \frac{\exp(\bm{z}_{i} \cdot \bm{z}_{j} / \tau) }{\sum_{k=1}^K  \exp(\bm{z}_{i} \cdot \bm{z}_{k} / \tau)}
\end{equation}
To leverage the class information from the anchors, we calculate the sum of the predicted probability distribution weighted by the similarity score to form a more semantically  informative target. We denotes this target as $p^{dc}_{i} \in R^{1 \times C}$, which can be expressed as the follows:
\begin{equation}
\label{equation:dctarget}
     p^{dc}_{i} = \displaystyle\sum_{j=1}^{K} p^{instance}_{i, j} \cdot p^{class}_{j}
\end{equation}

\begin{table*}[!h]
 \centering
\setlength\tabcolsep{2.5pt}
 \caption{Top-1 classification accuracy on ResNet-50 \cite{resnet} for CIFAR-10 and CIFAR-100 datasets. We compare cross-entropy, unsupervised representation learning (SimCLR \cite{simclr}), max-margin classifiers \cite{max-margin}, SupCon \cite{SupervisedCL}. We report the mean and std over 5 runs for CoNe. }
 \label{table:cifar}
\begin{tabular}{l c c c c c } 
\toprule 
DataSet   & SimCLR \cite{simclr}  & Cross-Entropy & Max-Margin \cite{max-margin} & SupCon & \bf{CoNe (Ours)} \\ \midrule
CIFAR-10 \cite{cifar}            & 93.6    & 95.0          & 92.4       & \bf{96.0}  & \bf{96.00}$\pm$0.11 \\
CIFAR-100 \cite{cifar}           & 70.7    & 75.3          & 70.5       & 76.5   & \bf{78.08}$\pm$0.09 \\ 
\bottomrule
\end{tabular}
\end{table*}

Now, we optimize the Kullback–Leibler divergence between $p^{class}_{i}$ and $p^{dc}_{i}$, which can be expressed by Eq. \eqref{equation:dcloss}. In this way, the features are constrained to have similar predicted probability distribution as their similar anchors, which further enhances the compactness.
\begin{equation}
\label{equation:dcloss}
     \mathcal{L}_{dc} = D_{KL} (p^{dc}_{i} \; || \; p^{class}_{i})
\end{equation}

\textbf{Overall Objective.} Finally, the overall training objective for CoNe will be Eq. \eqref{equation:overall}. where $\lambda_{sup}$ and $\lambda_{dc}$ are the balancing factors that control the weights of the two losses. Note that we only use $p_{class}$ as the prediction during the inference stage. 
\begin{equation}
\label{equation:overall}
     \mathcal{L}_{overall} = \mathcal{L}_{ce} + \lambda_{sup} \mathcal{L}^{sup}_{in}  + \lambda_{dc} \mathcal{L}_{dc}
\end{equation}

\section{Experiments}
\subsection{CIFAR-10 and CIFAR-100}
The CIFAR-10 dataset \cite{cifar} consists of 60000 32x32 color images in 10 classes, with 6000 images per class. There are 50000 training images and 10000 test images. CIFAR-100 \cite{cifar} is quite similar to the CIFAR-10; it has 100 categories containing 600 images in each class, and there are 500 training images and 100 testing images per class. 

\textbf{Implementations.} We adopt the ResNet-50 \cite{resnet} as our backbone encoder. Following the common practice for low-resolution datasets, we replace the first 7x7 Conv of stride 2 with 3x3 Conv of stride 1 and remove the first max pooling operation for ResNet-50. We attach a two-layer MLP head (with ReLU and BN in the hidden layer) to project the backbone feature from 2048-D into a 256-D space. Note that the additional computational cost of the projection head is negligible compared with the entire network, and we do not observe any improvements in terms of performance. The model is optimized for 1000 epochs by a standard SGD optimizer with a momentum of 0.9 and a weight decay of $0.0001$. We linearly warm up the learning rate for 100 epochs until it reaches 0.1 $\times$ BatchSize/256 (we use a batch size of 1024 by default), then switch to the cosine decay scheduler \cite{cosine_lr}. By following the setting in \cite{SupervisedCL}, we adopt the SimCLR based augmentation with color distortion (strength=0.5), and leave out Gaussian blur. For CoNe related hyper-parameters, we set $\lambda_{sup}=0.7, \lambda_{dc}=0.4, \tau_{sup}=0.1$, and $\tau_{dc}=0.07$. (Noted that we use $\tau_{sup}$ and $\tau_{dc}$ to denote the temperature parameter in Eq. \eqref{equation:supin} and Eq. \eqref{equation:instance_similarity} respectively.) We adopt a memory bank with 4096 past examples and use Top-32 features to compute $\mathcal{L}^{sup}_{in}$. The momentum coefficient is $m=0.996$ and increases to 1 with a cosine schedule.

\textbf{Performance on CIFAR Datasets}. The comparative performance between the proposed approach - CoNe, SimCLR \cite{simclr}, Cross-Entropy, Max-Margin \cite{max-margin}, and SupCon \cite{SupervisedCL} on CIFAR-10 and CIFAR-100 datasets is delineated in Table \ref{table:cifar}. As inferred from the results, both CoNe and SupCon demonstrate highly competitive performance on the CIFAR-10 dataset, attaining 96.0\%. However, with respect to the CIFAR-100 dataset, CoNe evidently outperforms SupCon, achieving a performance of 78.08\% which surpasses SupCon's 76.5\% by a large margin. 

\begin{table}[!h]
 \centering
 \vspace{-5mm}
 \caption{Comparison with PaCo \cite{paco} on CIFAR100 under stronger setting. The results for other methods are directly copied from PaCo \cite{paco}.}
 \label{table:paco_cifar100}
\begin{tabular}{l c c c} 
\toprule 
Objective & dataset & Epochs & Top-1 \\ \midrule
Cross-Entropy  & ResNet-50 &  400  & 77.9\\
Cross-Entropy + $\mathcal{L}^{sup}_{out}$  & ResNet-50 &  400  & 78.0\\
PaCo  & ResNet-50 &  400  & 79.1\\ \midrule
\bf{CoNe (Ours)}    & \bf{ResNet-50}  & \bf{400}  & \bf{81.2} \\ 
\bottomrule
\end{tabular}
\end{table}

\textbf{Comparing with PaCo under Stronger Setting}.  PaCo \cite{paco} employed a relatively advanced data augmentation methodology for a similar experiment. In order to conduct a fair comparison with PaCo, we utilized their official codebase\footnote{\url{https://github.com/dvlab-research/Parametric-Contrastive-Learning/blob/main/PaCo/LT/paco_cifar.py}} so as to align with their experimental conditions. The results, as illustrated in Table \ref{table:paco_cifar100}, demonstrate that our CoNe outperforms PaCo \cite{paco} by a margin of 1.1\% on this particular benchmark. This finding underscores the effective performance of CoNe.

\subsection{Experiments on ImageNet.}

\begin{table}[!h]
 \centering
 \setlength\tabcolsep{2.5pt}
 \vspace{-5mm}
 \caption{ Comparison with standard Cross Entropy under 100 Epochs Setting on ImageNet. For baseline, we report our reproduced results under the same training recipe.}
 \label{table:100epImageNet}
\begin{tabular}{l c c c c} 
\toprule 
Objective & Arch  & Aug & Epochs & Top-1 \\ \midrule
Cross-Entropy  & ResNet-18 & Standard & 100  & 70.7\\
\bf{CoNe (Ours)}    & \bf{ResNet-18} & \bf{Standard} & \bf{100}  & \bf{72.5} \\ \midrule
Cross-Entropy  & ResNet-34 & Standard & 100  & 74.3  \\
\bf{CoNe (Ours)}    & \bf{ResNet-34} & \bf{Standard} & \bf{100}  & \bf{75.4}  \\ \midrule
Cross-Entropy  & ResNet-50 & Standard & 100  & 76.9 \\
\bf{CoNe (Ours)}    & \bf{ResNet-50} & \bf{Standard} & \bf{100}  & \bf{78.7}  \\ \midrule
Cross-Entropy  & ResNet-101 & Standard & 100  & 78.7  \\
\bf{CoNe (Ours)}    & \bf{ResNet-101} & \bf{Standard} & \bf{100}  & \bf{79.6}  \\ \midrule
Cross-Entropy  & ResNet-152 & Standard & 100  & 79.4  \\
\bf{CoNe (Ours)}    & \bf{ResNet-152} & \bf{Standard} & \bf{100}  & \bf{80.3}  \\ \midrule
Cross-Entropy  & ResNet-200 & Standard & 100  & 79.6  \\ 
\bf{CoNe (Ours)}    & \bf{ResNet-200} & \bf{Standard} & \bf{100}  & \bf{80.5}  \\
\bottomrule
\end{tabular}
\end{table}

Subsequently, we implement an evaluation of our proposed methodologies on the ImageNet-1k dataset \cite{imagenet_cvpr09} to further substantiate the performance. The majority of hyper-parameters employed in this experiment are congruent with those used in our previous CIFAR tests, with the exception of our utilization of a larger memory bank, containing 65k past features, and construing Top-512 to compute $\mathcal{L}^{sup}_{in}$. In this experiment, different ResNet models were optimized over a period of 100 epochs, including a warm-up phase of 5 epochs, with standard data augmentation which only employs RandomResizedCrops and RandomFlip. The results are shown in Table \ref{table:100epImageNet}. For a fair comparison with the cross-entropy baseline, we report our reproduced results under exactly the same training strategy. Derived from the results, our proposed method significantly enhanced the classification accuracy by a margin of 0.9\% to 1.8\% across various ResNet architectures.

\begin{table}[h]
 \centering
 \setlength\tabcolsep{1.5pt}
 \vspace{-5mm}
 \caption{ Comparison of Cross-Entropy, SupCon \cite{SupervisedCL}, PaCo \cite{paco}, and CoNe under stronger augmentation and longer training epochs. The performance of Cross-Entropy and SupCon are directly copied from the \cite{SupervisedCL}. (w/ MB) denotes the memory buffer based implementation for SupCon.}
 \label{table:supcon}
\begin{tabular}{l c c c c c } 
\toprule 
Objective & Arch  & Aug & Epochs & Top-1 \\ \midrule
Cross-Entropy & ResNet-50 & MixUp         & 300  & 77.4 \\
Cross-Entropy & ResNet-50 & CutMix        & 300  & 78.6 \\
Cross-Entropy & ResNet-50 & AutoAug  & 300  & 78.2 \\
SupCon        & ResNet-50 & AutoAug  & 350  & 78.7 \\
SupCon (w/ MB) & ResNet-50 & AutoAug  & 350  & 79.1 \\
PaCo   & ResNet-50 & RandAug  & 400  & 79.3 \\
\bf{CoNe (Ours)} & \bf{ResNet-50} & \bf{AutoAug}  & \bf{350} & \bf{80.2} \\ \midrule

SupCon  & ResNet-101 & StackedAug  & 350  & 80.2 \\
PaCo  & ResNet-101 & StackedAug  & 400  & 80.9 \\
\bf{CoNe (Ours)}    & \bf{ResNet-101} & \bf{StackedAug}  & \bf{350}  & \bf{81.4} \\ \midrule
Cross-Entropy  & ResNet-200 & AutoAug     & 700  & 80.6 \\
Cross-Entropy  & ResNet-200 & StackedAug  & 700  & 80.9 \\
SupCon      & ResNet-200 & StackedAug  & 700  & 81.4 \\
PaCo  & ResNet-200 & StackedAug  & 400  & \textbf{81.8} \\
\bf{CoNe (Ours)}    & \bf{ResNet-200} & \bf{StackedAug}  & \bf{400}  & \bf{81.8} \\ 
\bottomrule
\end{tabular}
\end{table}

\textbf{Apple to Apple Comparison with SupCon.} 
We conducted an apple to apple of our proposed CoNe with SupCon under a fair condition. Specifically, we adhered to the conventional contrastive framework, augmenting each sample twice and forwarding both augmented perspectives through the encoder. We maintained congruence in augmentation strategies and training epochs as employed in the SupCon approach. The other experimental parameters, including temperature, weight decay, and more, were kept consistent with our earlier ImageNet experiment. The outcomes are presented in Table \ref{table:supcon}. Comparing the performance of CoNe and SupCon, our CoNe exhibits notable improvements of 1.1\% and 1.2\% for ResNet-50 and ResNet-101, respectively. Notably, when applied to ResNet-200, CoNe surpasses SupCon while demanding significantly fewer training epochs (400 Epochs vs. 700 Epochs). An important advantage of our approach is that CoNe operates as an end-to-end framework, eliminating the need for an additional 90 epochs of fine-tuning, which is essential in the case of SupCon. Moreover, CoNe demonstrates resilience with the standard SGD optimizer, in contrast to the sensitivity of SupCon to optimizer selection (SupCon employs LARS \cite{lars} and RMSProp \cite{rmsprop} for pre-training and fine-tuning stages). This robustness in optimization is a key characteristic of CoNe. Comparing with the recent PaCo method \cite{paco}, our CoNe achieves a competitive edge. For ResNet-50 and ResNet-101, CoNe attains 0.9\% and 0.5\% improvements with slightly fewer training epochs (350 Epochs vs. 400 Epochs), and the same results for ResNet-200, showcasing the inherent advantages of CoNe.

\begin{table}[h]
 \centering
 \setlength\tabcolsep{2pt}
 \vspace{-5mm}
 \caption{Compare with Timm strategy \cite{timmresnet}. Different from Table \ref{table:supcon}, the CoNe results in this table further includes three training strategies - label smoothing, exponential moving average, and train/inference resize tuning. }
 \label{table:timm}
\begin{tabular}{l c c c }
\toprule 
Objective                    & Arch        & Aug      & Top-1 \\ \midrule
Binary-Cross-Entropy    & ResNet-18   & Timm A1  & 71.5 \\
\bf{CoNe (Ours)}                    & \bf{ResNet-18}   & \bf{AutoAug}  & \bf{74.3} \\ \midrule
Binary-Cross-Entropy    & ResNet-34   & Timm A1  & 76.4 \\
\bf{CoNe (Ours)}                    & \bf{ResNet-34}   & \bf{AutoAug}  & \bf{78.0} \\ \midrule
Binary-Cross-Entropy    & ResNet-50   & Timm A1   & 80.4 \\
Cross-Entropy           & ResNet-50   & Timm B    & 79.4 \\
Cross-Entropy           & ResNet-50   & Timm C.1  & 79.8 \\
Cross-Entropy           & ResNet-50   & Timm C.2  & 80.0 \\
Binary-Cross-Entropy    & ResNet-50   & Timm D    & 79.8 \\
\bf{CoNe (Ours)}                    & \bf{ResNet-50}   & \bf{AutoAug}   & \bf{80.8}  \\ \midrule
Binary-Cross-Entropy    & ResNet-101  & Timm A1    & 81.5 \\
\bf{CoNe (Ours)}                    & \bf{ResNet-101}  & \bf{StackedAug} & \bf{82.1}  \\ \midrule
Binary-Cross-Entropy    & ResNet-152  & Timm A1   & 82.0 \\
\bf{CoNe (Ours)}                    & \bf{ResNet-152}  & \bf{StackedAug} & \bf{82.7} \\ \bottomrule
\end{tabular}
\end{table}

\textbf{Compare with State-of-the-arts.} To further explore the limits of our method, we try to incorporate CoNe with three commonly used training strategies. 1). Label Smoothing. 2) Exponential Moving Average, noted this is a different EMA network from our Eq. \eqref{equation:ema}. The difference is that Eq. \eqref{equation:ema} only averages the models parameters, but this EMA model also averages buffer information. The final performance will be tested on this EMA model. 3). Train/Inference resize tuning. We adjust the training resolution to $176\times176$ and inference resolution to $232\times232$ $\rightarrow$ center crop $224\times224$. We also reduce the weight decay to $6e-5$, and the rest of the settings are the same as in Table \ref{table:supcon}. In this experiment, we would like to compare CoNe with the existing state-of-the-art training recipe (Timm \cite{timmresnet}). The results are shown in Table \ref{table:timm}. Notably, our method surpasses Timm training recipe by a large margin.  We want to emphasize that the Timm training recipe \cite{timmresnet} adopts lots of advanced training strategies (\eg~ Stochastic Depth \cite{stochasticdepth}, CutMix \cite{cutmix}, MixUp\cite{mixup}, Repeated Augmentation \cite{reaptedaug1,reaptedaug2}, and LAMB optimizer \cite{lamb},). Incorporating more advanced training strategies might potentially improve results further. However, we need more sophisticated hyper-parameter tuning, which is not the focus of this paper. Since we have already achieved state-of-the-art performance, we would like to leave this problem as future work.

\begin{table*}
    \centering
    \setlength\tabcolsep{2.5pt}
    \caption{Transfer learning on downstream classification datasets. The performance of other methods are directly copied from \cite{SupervisedCL}. Following the same evaluation protocol from \cite{simclr, byol}, we report Top-1 accuracy except for Pets, Flowers, and Caltech101 for which we report mean per-class accuracy.}
    \begin{tabular}{l c c c c c c c c c c}
        \toprule
        Method & CIFAR10 & CIFAR100 &  Food & \; Cars \; & DTD & Pets & Flowers & Aircraft & Caltech101 &  Mean \\ \midrule
        SimCLR \cite{simclr} & 97.7 & 85.9 & 88.2 & 91.3 & 73.2 & 89.2 & 97.0 & \bf{88.1} & 92.1 & 89.2 \\
        CE \cite{simclr} &  96.5 & 85.1 & 87.4 & 89.6 & 76.9 & 92.4 & 96.9 & 80.8 & 92.3 & 88.7 \\ 
        SupCon &  97.4 & 84.3 & 87.2 & 91.7 & 74.6 & 93.5 & 96.0 & 84.1 & 91.0 & 88.9 \\ \midrule
        \bf{CoNe} & \bf{97.9} & \bf{86.2} & \bf{88.5} & \bf{91.9} & \bf{77.6} & \bf{94.7} & \bf{97.7} & 88.0 & \bf{94.4} & \bf{90.8} \\ \bottomrule
    \end{tabular}
    \label{table:transfer_classification}
\end{table*}

\begin{table*}[h]
 \centering
 \setlength\tabcolsep{4pt}
 \caption{Transfer learning on object detection and instance segmentation.}
 \label{table:transfer_detection}
\begin{tabular}{l c c c  c c c } 
\toprule 
& \multicolumn{3}{c}{COCO detection} & \multicolumn{3}{c}{COCO instance seg.} \\
\cmidrule(l{3pt}r{3pt}){1-1} \cmidrule(l{3pt}r{3pt}){2-4}  \cmidrule(l{3pt}r{3pt}){5-7}
Method   &  AP$_{50}^{Box}$ & AP$^{Box}$ & AP$_{75}^{Box}$ & AP$_{50}^{Mask}$ &  AP$^{Mask}$ & AP$_{75}^{Mask}$ \\ \midrule
\emph{ResNet-50 with C4 Backbone} \\ 
 Cross-Entropy (Supervised)  &  58.2 & 38.2 & 41.2 & 54.7 & 33.3 & 35.2 \\
 SimCLR \cite{simclr} & 57.7 & 37.9 & 40.9 & 54.6 & 33.3 & 35.3 \\
 MoCo v2 \cite{mocov2} & 58.8 & 39.2 & 42.5 & 55.5 & 34.3 & 36.6 \\
 SwAV \cite{swav} & 58.6 & 38.4 & 41.3 & 55.2 & 33.8 & 35.9 \\
 SimSiam \cite{SimSiam} & 59.3 & 39.2 & 42.1 & 56.0 & 34.4 & 36.7 \\ 
 Barlow Twins \cite{barlowtwins} & 59.0 & 39.2 & 42.5 & 56.0 & 34.3 & 36.5 \\ 
 \bf{CoNe (ours)} & \bf{59.9} & \bf{39.8} & \bf{42.9} & \bf{56.3} & \bf{34.6} & \bf{36.8} \\ \midrule
 \emph{ResNet-50 with FPN} \\ 
 Cross-Entropy (Supervised) \cite{sl-mlp}  & 61.1 & 40.1 & 43.8 & 57.7 & 35.7 & 38.0 \\
 Cross-Entropy w/ CutMix (Supervised) \cite{cutmix}  & 60.9 & 40.8 & 44.3 & 57.8 & 36.8 & 39.5 \\
 SupCon (Supervised) \cite{SupervisedCL}  & 61.2 & 41.0 & 44.7 & 58.2 & 37.0 & 39.6 \\
 SL-MLP (Supervised) \cite{sl-mlp}  & \bf{61.8} & 40.7 & 44.2 & 58.4 & 36.1 & 38.5 \\
 \bf{CoNe (Ours)}  & 61.5 & \bf{41.1} & \bf{45.1} & \bf{58.5} & \bf{37.2} & \bf{40.1} \\
 \bottomrule
\end{tabular}
\end{table*}

\begin{table}[!h]
 \centering
 \setlength\tabcolsep{2pt}
 \caption{More Experiments with Different Architectures.}
 \label{table:light}
\begin{tabular}{l c c c} 
\toprule 
Objective & Arch & Epochs & Top-1 \\ \midrule
Cross-Entropy & RegNetX-400MF & 100  & 72.8\\
\bf{CoNe (Ours)}   & \bf{RegNetX-400MF}  & \bf{100}  & \bf{74.6} \\ \midrule
Cross-Entropy & MobileNet V2  & 150  & 71.9\\
\bf{CoNe (Ours)}   & \bf{MobileNet V2}  & \bf{150}  & \bf{73.8}  \\ \midrule
Cross-Entropy & ShuffleNet v2 1.0$\times$  & 240  & 69.4 \\
\bf{CoNe (Ours)}   & \bf{ShuffleNet v2 1.0$\times$}  & \bf{240}  & \bf{72.2}  \\ \midrule
Cross-Entropy & DeiT-Tiny  & 300  & 74.4 \\ 
\bf{CoNe (Ours)}   & \bf{DeiT-Tiny}  & \bf{300}  & \bf{76.0}  \\
\bottomrule
\end{tabular}
\end{table}

\textbf{More Experiments with Different Architectures}
We further demonstrate the generality of CoNe by training it with various architectures. Specifically, we implement CoNe with RegNet \cite{regnet}, MobileNet v2 \cite{mobilenetv2}, ShuffleNet v2 \cite{shufflenetv2}, and Vision Transformer \cite{vit, deit}. From Table \ref{table:light}, we can see that CoNe consistently improves the performance across different architectures.

\textbf{Transfer Learning on Classification Tasks.} We also show the transferability of CoNe on various downstream classification datasets. This experiment adopts the pre-trained ResNet-50 model from Table \ref{table:supcon}. Concretely, we fine-tune our pre-trained ResNet-50 network on CIFAR-10 \cite{cifar}, CIFAR-100 \cite{cifar}, Food101 \cite{food101}, Cars \cite{cars}, DTD \cite{dtd}, Oxford-IIIT-Pets \cite{pets}, Aircrat \cite{aircraft}, Oxford-Flowers \cite{flowers}, and Caltech-101 \cite{caltech101}. The results are shown in Table \ref{table:transfer_classification}; our CoNe achieves the best performance on \textbf{8 out of 9} datasets.

\textbf{Transfer Learning on Detection and Segmentation.} Next, we evaluate the representation quality by transferring the model to object detection and instance segmentation tasks on the COCO dataset \cite{coco}. Since a lot of recent unsupervised pretraining methods claim their performance has surpassed the supervised pretrain. We would like to compare our method with both unsupervised and supervised methods. Specifically, the CoNe pre-trained parameters (from Table \ref{table:supcon}) will serve as the initialization for Mask-RCNN \cite{maskrcnn} with C4 and FPN backbone. We would like to compare the performance by following the configuration in \cite{moco}, and \cite{goodview}. We fine-tune the model on the \emph{train2017} set and evaluate on \emph{val2017}. The schedule is the default 1x in Detectron2 \cite{detectron2}. We report the standard evaluation metric AP$_{50}$, AP, and AP$_{75}$ for detection and segmentation. Table \ref{table:transfer_detection} shows that CoNe surpasses prior arts (\eg~SimSiam and SL-MLP) on these localization-based tasks and is significantly better than the cross-entropy baseline.

\begin{table}[h]
 \centering
 \setlength\tabcolsep{2pt}
 \caption{ Robustness experiments on ImageNet-C. Note that we report the mCE in this experiment; \textbf{lower mCE indicates better performance.} }
 \label{table:robust}
\begin{tabular}{l c c c }
        \toprule
        Arch & Cross-Entropy & SupCon & \textbf{CoNe (Ours)}  \\ \midrule
        ResNet-50 & 68.6 & 67.2 & \bf{52.7}  \\
        ResNet-200 & 52.4 & 50.6 & \bf{39.9}  \\\bottomrule
    \end{tabular}
\end{table}

\textbf{Robustness to Image Corruptions}
Next, we also test the robustness of our CoNe on the ImageNet-C dataset \cite{imagenetc}, which consists of 15 types of natural corruptions. Followed by the standard benchmark in \cite{imagenetc}, we adopt the Mean Corruption Error (mCE) as our metric. The results are shown in Table \ref{table:robust} below. As we can observe that CoNe dramatically improves the robustness of the model.

\section{Ablation study}
In this section, we will empirically study our CoNe based on various conditions and show the effect of each component and hyper-parameter sensitivities of our methods. For all experiments in this section, we adopt the ResNet-50 as our backbone and train the model on ImageNet for 100 epochs. We use the most standard augmentation policies (\ie RandomResizedCrops and RandomFlip) by default unless we have mentioned.

\begin{table}[h]
 \centering
 \setlength\tabcolsep{8pt}
 \caption{Effect of each component in our proposed method.}
 \label{table:effect}
\begin{tabular}{c c c c c c}
\toprule 
$\mathcal{L}_{ce}$   &  Proj & $\mathcal{L}^{sup}_{in}$ & $\mathcal{L}^{sup}_{out}$ & $\mathcal{L}_{dc}$ &  Top-1 \\ \midrule
 \checkmark & & & & & 76.9 \\
 \checkmark & \checkmark & & & & 76.8 \\
 \checkmark & \checkmark & \checkmark & &  & 78.1 \\
 \checkmark & \checkmark & & \checkmark &  & 76.9 \\ \midrule
 \checkmark & \checkmark & \checkmark & &  \checkmark & \bf{78.7} \\
\bottomrule
\end{tabular}
\end{table}

\textbf{Effect of Each Components}. We first show the effect of each loss function in Table \ref{table:effect}. The ResNet-50 achieves 76.9\% Top-1 accuracy with the vanilla cross-entropy loss, which is our baseline. Adding a projection head does not affect the performance ($2^{nd}$ row); we want to emphasize again that this projection head is only used to reduce the dimension since we need to store a large number of features in the memory bank; it just introduces an additional 7M FLOPS, which is negligible compared with the entire network (4.1G). Next, we try to joint train the model with $\mathcal{L}^{sup}_{in}$ loss as Eq. \eqref{equation:cone}. The results in $3^{rd}$ row show that $\mathcal{L}^{sup}_{in}$ substantially improve the baseline by +1.2\%. We also tried to replace $\mathcal{L}^{sup}_{in}$ by $\mathcal{L}^{sup}_{out}$ and performed an extensive hyper-parameter search on the temperature and loss weight ($4^{th}$ row). However, the best result we can get is 76.9\% which has no difference from our baseline. Finally, further incorporating the distributional consistency loss could bring +0.6\% improvements.

\begin{table}[h]
 \centering
 \caption{Effect of including/excluding contrastive view for $\mathcal{L}^{sup}_{in}$.}
 \label{table:contra_view}
\begin{tabular}{l c c c}
\toprule 
Augmentation & W/ Contra & W/o Contra & Diff \\ \midrule
Standard      & 77.5 & 78.1 & +0.6 \\
AutoAug \cite{autoaugment}   & 78.1 & 78.3 & +0.2 \\ 
RandAug  \cite{randaugment}    & 77.9 & 78.1 & +0.2 \\
SimCLR Aug \cite{simclr}   & 76.9 &  77.2 & +0.3 \\  \bottomrule
\end{tabular}
\end{table}

\textbf{Effect of Contrastive View}. We also conduct experiments to ablate the effect of including/excluding the contrastive view in $\mathcal{L}^{sup}_{in}$. We tried both strong (AutoAug \cite{autoaugment}, RandAug \cite{randaugment}, SimCLR Aug \cite{simclr}) and weak augmentations in this experiment. We do not include the $\mathcal{L}_{dc}$ and train the model for 100 epochs. The results are in Table \ref{table:contra_view}. Basically, when the standard augmentation is adopted, excluding the contrastive view is +0.6\% better than including it. When we work with a stronger augmentation, the performance gaps between these two settings can be reduced, but the excluding the contrastive view is consistently slightly better than including it. Thus, we would exclude the contrastive view by default. We believe such experiments should further support our analysis in Section 3.3.

\begin{table}[h]
 \centering
 \caption{Comparing with Other Nearest Neighbor Contrast Method.}
 \label{table:other_nn}
\begin{tabular}{l c c c}
\toprule 
Method & Cross-Entropy & Epoch & Top-1 \\ \midrule
baseline   & &  100 & 76.9  \\
CMSF \cite{cmsf}   &  &  200 & 76.4  \\
CMSF \cite{cmsf}   & \checkmark &  100 & 77.0  \\
NNCLR \cite{nnclr}   & \checkmark &  100 & 77.5  \\
\textbf{CoNe (Ours)}  & \checkmark & \textbf{100} & \textbf{78.1} \\
\bottomrule
\end{tabular}
\end{table}

\textbf{Comparing with Other Nearest Neighbor Contrast Method}. 
The concept of the nearest neighbor contrast method has been introduced in previous research \cite{cmsf, nnclr}. However, our approach, CoNe, stands apart from the methodologies presented in \cite{cmsf, nnclr} due to its fundamental differences. In comparison to \cite{cmsf}, our CoNe method offers a distinct characteristic – the capability to yield a more substantial gradient for similar positive instances. This distinguishing feature holds immense significance and has been a key focus in our work. Furthermore, when comparing with \cite{nnclr}, it becomes evident that the latter exclusively attracts the 1-nearest neighbor (1-NN), while neglecting other positive samples. In contrast, our CoNe method accounts for a broader spectrum of positive instances, resulting in a more comprehensive learning process. To substantiate the efficacy of CoNe, we conducted an experiment comparing its performance against that of \cite{cmsf} and \cite{nnclr}. The results of this experiment are presented in Table \ref{table:other_nn}. The outcomes unmistakably demonstrate the superior performance of our CoNe method compared to other nearest neighbor contrast techniques in a supervised setting.

\begin{table}[h]
    \centering
    \setlength\tabcolsep{10pt}
    \vspace{-8mm}
    \caption{Hyper-Parameter Sensitivity for $\lambda_{sup}$ and $\tau_{sup}$ }
    \begin{tabular}{l | c c c c}
    \diagbox{$\lambda_{sup}$}{$\tau_{sup}$} & 0.05 & 0.07 & 0.1 & 0.2 \\ \hline
    0.3 & 77.5 & 77.4 & 77.6 & 77.3 \\ 
    0.7 & 77.6 & 78.0 & \bf{78.1} & 78.1 \\ 
    1.0 & 77.7 & 78.0 & \bf{78.1} & 78.0 \\
    \end{tabular}
    \label{table:ablation_tq}
    \vspace{-12mm}
\end{table}

\begin{table}[h]
    \centering
    \setlength\tabcolsep{10pt}
    \caption{Hyper-Parameter Sensitivity for $\lambda_{dc}$ and $\tau_{dc}$}
    \begin{tabular}{l | c c c c}
    \diagbox{$\lambda_{dc}$}{$\tau_{dc}$} & 0.05 & 0.07 & 0.1 & 0.2 \\ \hline
    0.2 & 78.6 & 78.4 & 78.6 & 78.5 \\ 
    0.4 & 78.5 & \bf{78.7} & \bf{78.7} & 78.4 \\ 
    0.6 & 78.5 & \bf{78.7} & 78.5 & 78.4 \\
    \end{tabular}
    \label{table:ablation_tk}
\end{table}

\begin{figure*}[h]
    \centering
    \includegraphics[width=\linewidth]{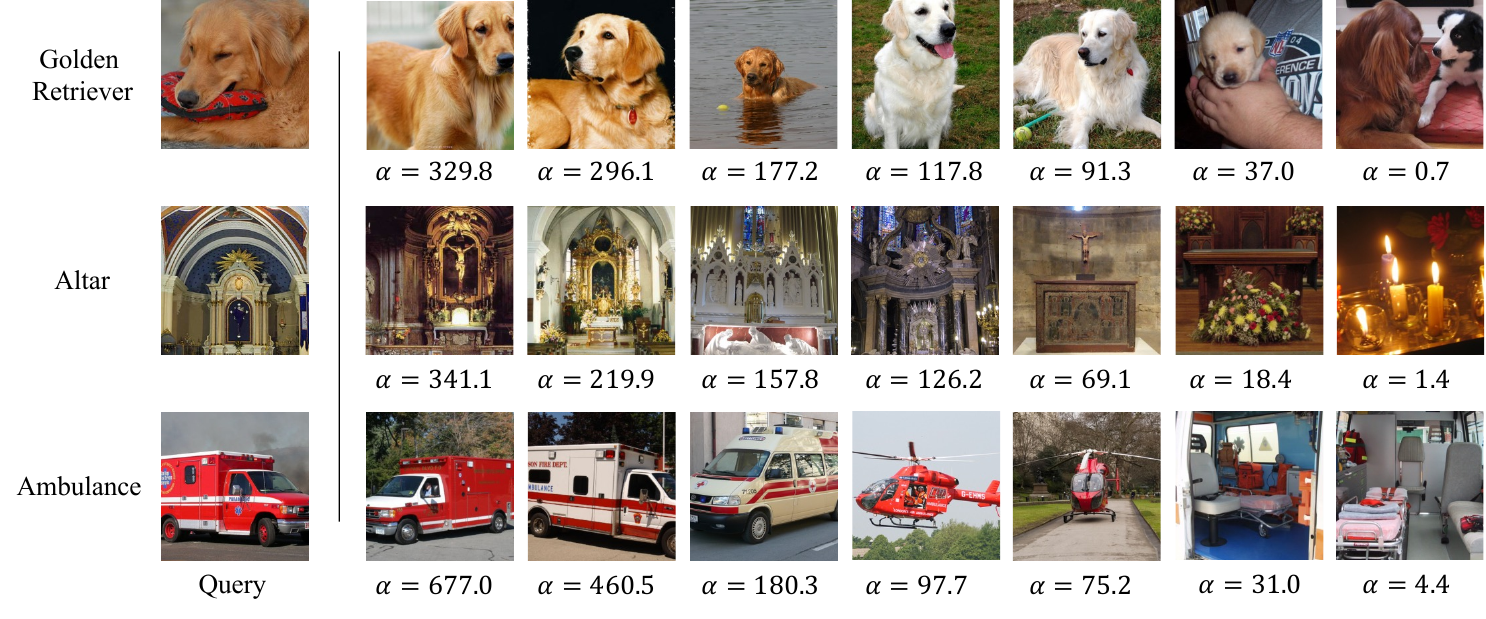}
    \caption{Visualization results of the gradient coefficient $\alpha$ in Eq. \eqref{equation:gradeint}}
    \label{fig:examples_coefficient}
\end{figure*}

\textbf{Temperature and Weight for $\mathcal{L}^{sup}_{in}$ and $\mathcal{L}_{dc}$}. To find the best configuration for $\mathcal{L}^{sup}_{in}$, we do not include $\mathcal{L}_{dc}$ in this experiment. Specifically, we sweep over $[0.3, 0.7, 1.0]$ for $\lambda_{sup}$ and $[0.05, 0.07, 0.1, 0.2]$ for $\tau_{sup}$. The results are shown in Table \ref{table:ablation_tq}. Basically, an arbitrary setting could significantly improve the baseline, We use $\lambda_{sup}=0.7$ and $\tau_{sup}=0.1$ as our default setting. To study the influence of hyper-parameters for $\mathcal{L}_{dc}$, we simply fix the optimal value for $\mathcal{L}^{sup}_{in}$, we also sweep over $[0.2, 0.4, 0.6]$ for $\lambda_{dc}$ and $[0.05, 0.07, 0.1, 0.2]$ for $\tau_{dc}$. As can be seen in Table \ref{table:ablation_tk}, $\mathcal{L}_{dc}$ can always improve the performance, We use $\lambda_{dc}=0.4$ and $\tau_{dc}=0.07$ as our default setting.

\begin{table}[h]
    \centering
    \setlength\tabcolsep{10pt}
    \caption{Hyper-Parameter Sensitivity for Momentum}
    \begin{tabular}{l | c c c c}
    Momentum & 0.98 & 0.99 & 0.996 & 0.999 \\ \hline
    Top-1    & 78.5 & 78.6 & \bf{78.7} & 78.5 \\ 
    \end{tabular}
    \label{table:ablation_m}
    \vspace{-10mm}
\end{table}

\begin{table}[h]
    \centering
    \setlength\tabcolsep{10pt}
    \caption{\footnotesize Hyper-Parameter Sensitivity for Top-N}
    \begin{tabular}{l | c c c c}
    Top-N  & 128 & 256 & 512 & 1024 \\ \hline
    Top-1  & \bf{78.7} & \bf{78.7} & \bf{78.7} & 78.5 \\ 
    \end{tabular}
    \label{table:ablation_n}
    \vspace{-10mm}
\end{table}

\begin{table}[h]
    \centering
    \setlength\tabcolsep{10pt}
    \caption{\footnotesize Hyper-Parameter Sensitivity for Memory Buffer Size}
    \begin{tabular}{l | c c c c}
    Buffer Size  & 16K & 65K & 130K & 260K \\ \hline
    Top-1  & 78.6 & \bf{78.7} & \bf{78.7}  & 78.6 \\
    \end{tabular}
    \label{table:ablation_k}
\end{table}

\textbf{EMA Related Hyper-parameters}. Next, we would like to examine the effect of various EMA-related hyper-parameters. Note that we directly adopt the optimal value for $\lambda_{sup}$, $\lambda_{dc}$, $\tau_{sup}$, and $\tau_{dc}$ in this experiment. Concretely, we study the consequence of memory bank size, the number of Top-$N$ for computing $\mathcal{L}^{sup}_{in}$, and the momentum updated coefficient value. We have shown the performance for these factors under different settings in Table \ref{table:ablation_m} \ref{table:ablation_n} \ref{table:ablation_k}. CoNe is quite robust against these hyper-parameters; the worst result is only about 0.2\% lower than the best setting.

\textbf{Visualizations for Gradient Coefficient}. To further verify our analysis for the gradient of $\mathcal{L}^{sup}_{in}$, we randomly select some images and calculate the gradient coefficient $\alpha$ with respect to their positive samples. We present the visualization results in Figure \ref{fig:examples_coefficient}. Obviously, a more semantic similar sample will result in a larger $\alpha$, which contributes more to the optimization direction. 

\begin{figure}[h]
    \centering
    \includegraphics[width=\linewidth]{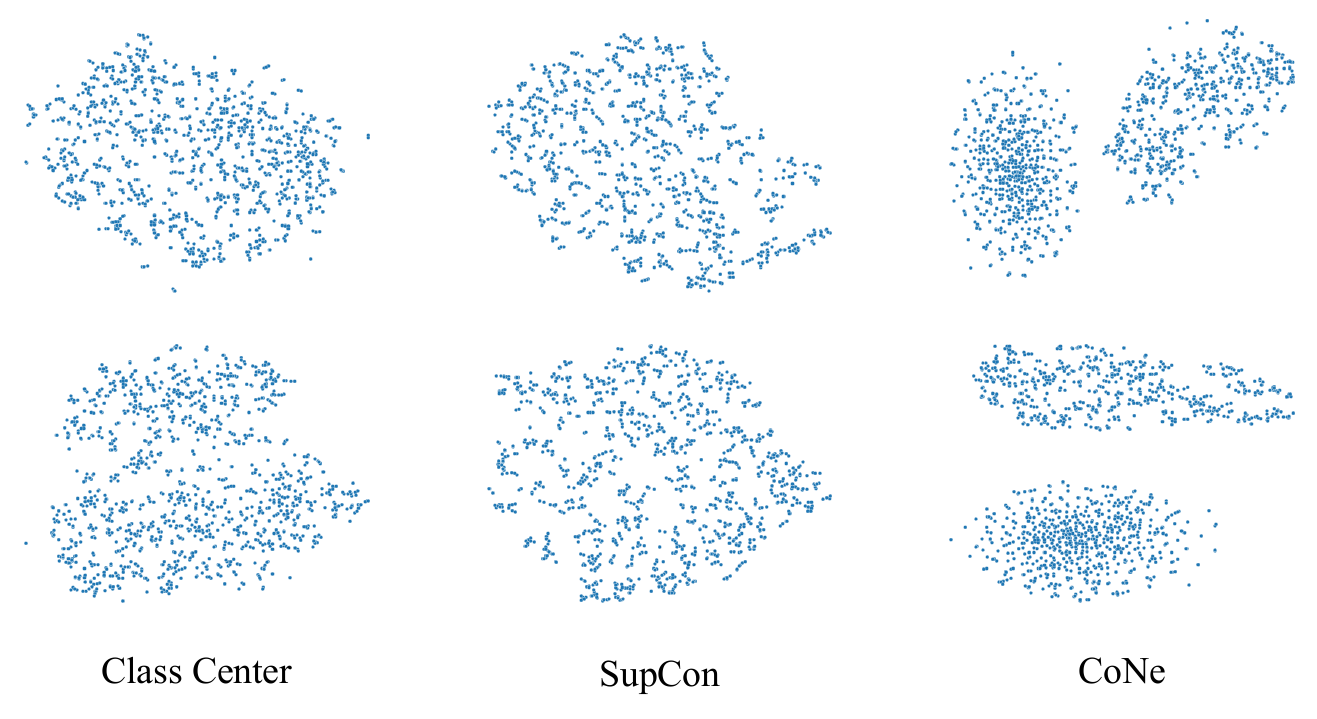}
    \caption{t-SNE visualization for the class center-based method, SupCon, and our CoNe on intra-class samples.}
    \label{fig:tsne}
\end{figure}
\textbf{t-SNE Visualizations}. Finally, we perform a t-SNE visualization \cite{tsne} for the class center-based method, SupCon, and our CoNe on intra-class samples. We randomly select two classes from the ImageNet and present the visualization results in Figure \ref{fig:tsne}.  Note that the features shown in each figure belong to the same class. As can be seen, the class center-based method and SupCon does not consider the discrepancy among intra-class instances since the intra-class features are randomly scattered across the figure, whereas our CoNe has a clear separation among the intra-class features. We believe this visualization results could show the characteristics that we present in Figure \ref{fig:framework}.

\section{Conclusion}
In this work, we propose Contrast with Your Neighbours (CoNe), a new supervised learning framework for image classification which utilizes $\mathcal{L}^{sup}_{in}$ in a more proper way. Theoretical analysis shows that $\mathcal{L}^{sup}_{in}$ loss aims to pull the feature of the training samples toward its similar positives, and more similar anchor will have a greater contribution to the optimization direction. Incorporating with classical cross-entropy loss significantly enhance the power of $\mathcal{L}^{sup}_{in}$. We also introduce the distributional consistency, which further enhances the compactness of similar samples and improves the classification performance. An extensive empirical study shows the effect of each component in our framework. The experiments on large-scale datasets and various architectures demonstrate the state-of-the-art performance for image classification problems. The current limitation of this work is that we have not tried CoNe to cooperate with the most advanced training recipe (\eg~\cite{timmresnet}); since it requires more experiments to find the optimal setting; we will leave this problem as our future work. 

\bibliography{sn-bibliography}%

\onecolumn
\begin{appendices}
\section{More details on Eq(3)}
\begin{equation}
\begin{aligned}
\footnotesize
    \mathcal{L}^{sup}_{in} &= - \log \frac{\displaystyle\sum_{p \in Pos(i)}  \exp(\bm{z}_{i} \cdot \bm{z}_{p} / \tau) }{\displaystyle\sum_{p \in Pos(i)}  \exp(\bm{z}_{i} \cdot \bm{z}_{p} / \tau) + \displaystyle\sum_{n \in Neg(i)}  \exp(\bm{z}_{i} \cdot \bm{z}_{n} / \tau)} \\
    &= \log \Bigg[ 1 + \frac{\displaystyle\sum_{n \in Neg(i)}  \exp(\bm{z}_{i} \cdot \bm{z}_{n} / \tau)}{\displaystyle\sum_{p \in Pos(i)}  \exp(\bm{z}_{i} \cdot \bm{z}_{p} / \tau) } \Bigg] \\
    &= \log \Bigg[ 1 + \frac{ \exp  \Big( \log \displaystyle\sum_{n \in Neg(i)}  \exp(\bm{z}_{i} \cdot \bm{z}_{n} / \tau) \Big) }{ \exp \Big( \log \displaystyle\sum_{p \in Pos(i)}  \exp(\bm{z}_{i} \cdot \bm{z}_{p} / \tau) \Big) } \Bigg] \\
    &= \log \Bigg[ 1 + \exp(\log \sum_{n \in Neg(i)} \exp(\bm{z}_{i} \cdot \bm{z}_{n} / \tau) - \log \sum_{p \in Pos(i)} \exp(\bm{z}_{i} \cdot \bm{z}_{p} / \tau) ) \Bigg] \\
\end{aligned}
\end{equation}

\noindent Since
\begin{equation}
\begin{aligned}
\footnotesize
\log \displaystyle\sum_{i}^{N} \exp(\bm{x}_i) = \max(\bm{x})  + \log \displaystyle\sum_{i}^{N} \exp(\bm{x}_i - \max(\bm{x}))
\end{aligned}
\end{equation}
Thus, the approximation error for $LogSumExp$ is Eq. \eqref{equation:logsumexp_error}. As we can see, the value of the error term is mainly affected by 1) the number of elements that are close to $\max(\bm{x})$, (i.e. when the elements in $TopK(\bm{x})$ are similar to each other, the bias would be large). 2) the total number of elements.
\begin{equation}
\begin{aligned}
\footnotesize
    m = \log \displaystyle\sum_{i}^{N} \exp(\bm{x}_i - \max(\bm{x}))
\end{aligned}
\label{equation:logsumexp_error}
\end{equation}
Finally, the objective of $\mathcal{L}^{sup}_{in}$ can be derived as Eq. \eqref{equation:derive_objective}, where $m_{n}$ and $m_{p}$ are the approximation error for positive anchors and negative anchors respectively.
\begin{equation}
\begin{aligned}
\footnotesize
\max(\bm{z}_{i} \cdot \bm{z}_{n}) - \max(\bm{z}_{i} \cdot \bm{z}_{p}) + m_{n} - m_{p}
\end{aligned}
\label{equation:derive_objective}
\end{equation}
We perform a numerical experiment to test the actual value for positive and negative anchors. The averaged $m_{p}$ for positive similarities is around [0.06, 0.1] across the training process. Considering the range of cosine similarity [-1, 1], an error term less than 0.1 should be a reasonable approximation. For negative, the $m_{n}$ is around [0.3, 0.45], which is much larger.

In this way, the overall error term $(m_{n} - m_{p})$ actually acts as a margin (just like the margin in triplet loss) to further enhance the separability between $\max(\bm{z}_{i} \cdot \bm{z}_{n})$ and $\max(\bm{z}_{i} \cdot \bm{z}_{p})$. We want to emphasize that this margin is not the key point of our paper; we mainly focus on how to deal with the positives. It will not affect our conclusion since improving the separability between $\max(\bm{z}_{i} \cdot \bm{z}_{n})$ and $\max(\bm{z}_{i} \cdot \bm{z}_{p})$ still can not guarantee that all $\bm{z}_{i} \cdot \bm{z}_{p}$ will be larger than $\max(\bm{z}_{i} \cdot \bm{z}_{n})$.

\section{More details on Eq(4)}
We first define $A(i)$ as the union set of $Pos(i)$ and $Neg(i)$. Then, the gradient of $\mathcal{L}^{sup}_{in}$ can be derived:
\begin{equation}
\begin{aligned}
\footnotesize
- \frac{ \partial \mathcal{L}^{sup}_{in} }{ \partial \, \bm{z}_i }  &= \frac{\partial }{ \partial \bm{z}_i} \log \frac{\displaystyle\sum_{p \in Pos(i)}  \exp(\bm{z}_{i} \cdot \bm{z}_{p} / \tau) }{\displaystyle\sum_{a \in A(i)}  \exp(\bm{z}_{i} \cdot \bm{z}_{a} / \tau)} \\ 
&= \frac{\partial }{\partial \bm{z}_i} \log \sum_{p \in Pos(i)} \exp(\bm{z}_{i} \cdot \bm{z}_{p} / \tau) - \frac{\partial }{\partial \bm{z}_i} \log \sum_{a \in A(i)} \exp(\bm{z}_{i} \cdot \bm{z}_{a} / \tau) \\
&= \frac{1}{\tau} \frac{\displaystyle\sum_{p \in Pos(i)} \bm{z}_{p} \exp(\bm{z}_{i} \cdot \bm{z}_{p} / \tau) }{\displaystyle\sum_{p \in Pos(i)}  \exp(\bm{z}_{i} \cdot \bm{z}_{p} / \tau)} - \frac{1}{\tau} \frac{\displaystyle\sum_{A \in A(i)} \bm{z}_{a} \exp(\bm{z}_{i} \cdot \bm{z}_{a} / \tau) }{\displaystyle\sum_{a \in A(i)}  \exp(\bm{z}_{i} \cdot \bm{z}_{a} / \tau)} \\
&= \frac{1}{\tau} \Bigg[ \frac{\displaystyle\sum_{p \in Pos(i)} \bm{z}_{p} \exp(\bm{z}_{i} \cdot \bm{z}_{p} / \tau) }{\displaystyle\sum_{p \in Pos(i)}  \exp(\bm{z}_{i} \cdot \bm{z}_{p} / \tau)} -  \frac{\displaystyle\sum_{p \in Pos(i)} \bm{z}_{p} \exp(\bm{z}_{i} \cdot \bm{z}_{p} / \tau) }{\displaystyle\sum_{a \in A(i)}  \exp(\bm{z}_{i} \cdot \bm{z}_{a} / \tau)} - \frac{\displaystyle\sum_{n \in Neg(i)} \bm{z}_{n} \exp(\bm{z}_{i} \cdot \bm{z}_{n} / \tau) }{\displaystyle\sum_{a \in A(i)}  \exp(\bm{z}_{i} \cdot \bm{z}_{a} / \tau)}  \Bigg] \\
&= \frac{1}{\tau} \Bigg[ \displaystyle\sum_{p \in Pos(i)} \bm{z}_p \underbrace{\Big[ \frac{\exp(\bm{z}_{i} \cdot \bm{z}_{p} / \tau)}{S_p} - \frac{\exp(\bm{z}_{i} \cdot \bm{z}_{p} / \tau)}{S_p + S_n} \Big] }_{\text{Coefficient $\alpha_p$}} + \displaystyle\sum_{n \in Neg(i)} \bm{z}_n \underbrace{ - \frac{\exp(\bm{z}_{i} \cdot \bm{z}_{n} / \tau)}{S_p + S_n}}_{\text{Coefficient $\alpha_p$}} \Bigg]
\end{aligned}
\end{equation}
where we have
\begin{align}
\footnotesize
    \mathcal{S}_p  &= \sum_{p \in Pos(i)} \exp(\bm{z}_{i} \cdot \bm{z}_{p} / \tau)   &   \mathcal{S}_n  &= \sum_{n \in Neg(i)} \exp(\bm{z}_{i} \cdot \bm{z}_{n} / \tau )
\end{align}

The $\alpha_p$ and $\alpha_n$ indicate the coefficient value for the gradient of the positive anchor and negative anchor, respectively. Note that the negative coefficient value $\alpha_n$ of $\mathcal{L}_{sup}^{in}$ and $\mathcal{L}_{sup}^{out}$ is the same. Thus, we do not have too many discussions on it. The property of the positive gradient has been analyzed in Section 3.2 of the main content of this paper.
\end{appendices}

\end{document}